\documentclass[sigconf, nonacm]{acmart}

\AtBeginDocument{%
  \providecommand\BibTeX{{%
    Bib\TeX}}}

\usepackage{array}
\usepackage{graphicx} % Required for \resizebox
\newcolumntype{P}[1]{>{\centering\arraybackslash}p{#1}}
\usepackage{makecell}
\usepackage{enumitem}
\usepackage{multicol}
\usepackage{multirow}
\usepackage{titlesec}

\newcommand{\BibTeX}{\rm B\kern-.05em{\sc i\kern-.025em b}\kern-.08em\TeX}

\renewcommand\footnotetextcopyrightpermission[1]{} % Removes the copyright box at the bottom

\begin{document}

\title{Label-free Industrial Fault Detection via Adversarial Inverse Reinforcement Learning: A System for Run-to-Failure Prognostics}

% \author{Dhiraj Neupane, Mohamed Reda Bouadjenek, Richard Dazeley, Sunil Aryal}
% \affiliation{
%   \institution{Deakin University}
%   \city{Waurn Ponds, Victoria}
%   \country{Australia}}
%   %\orcid{0000-0001-6548-311X}
% \email{{d.neupane,reda.bouadjenek,richard.dazeley,sunil.aryal}@deakin.edu.au}

\author{Dhiraj Neupane}
\affiliation{
  \institution{Deakin University}
  \city{Waurn Ponds, Victoria}
  \country{Australia}}
  \orcid{0000-0001-6548-311X}
\email{s222439738@deakin.edu.au}
\author{Mohamed Reda Bouadjenek}
\affiliation{
  \institution{Deakin University}
  \city{Waurn Ponds, Victoria}
  \country{Australia}}
  \orcid{0000-0003-1807-430X}
\email{reda.bouadjenek@deakin.edu.au}
\author{Richard Dazeley}
\affiliation{
  \institution{Deakin University}
  \city{Waurn Ponds, Victoria}
  \country{Australia}}
  \orcid{0000-0002-6199-9685}
\email{richard.dazeley@deakin.edu.au}
\author{Sunil Aryal}
\affiliation{
  \institution{Deakin University}
  \city{Waurn Ponds, Victoria}
  \country{Australia}}
  \orcid{0000-0002-6639-6824}
\email{sunil.aryal@deakin.edu.au}

\renewcommand{\shortauthors}{Neupane et al.}

\begin{abstract}
Machinery fault detection (MFD) remains heavily reliant on supervised learning, which struggles with the scarcity of fault labels in real-world settings. While reinforcement learning (RL) offers a framework to model the sequential nature of degradation, current ``RL-based'' MFD methods reduce the problem to a static contextual bandit (CB) formulation: by ignoring state transitions and discarding the temporal discount factor, they collapse to standard supervised classification.
We propose an adversarial inverse reinforcement learning (AIRL) framework that treats MFD as an offline IRL problem. Unlike reconstruction-based approaches that rely on static error margins, or CBs that ignore dynamics, our method recovers an intrinsic ``health'' reward directly from observational state transitions, requiring neither manual reward engineering nor fault labels. On three run-to-failure benchmarks (HUMS2023, IMS, XJTU-SY), AIRL is the only method achieving non-saturated post-detection consistency across all datasets, while CB baselines fail to detect gradual degradation and reconstruction models collapse into always-anomalous states. Code and data: \href{https://github.com/dhirajneupane/AIRL-MFD-DN}{https://github.com/dhirajneupane/AIRL-MFD-DN}.
\end{abstract}


% \ccsdesc[500]{Reinforcement learning~Adversarial Inverse Reinforcement learning}
% \ccsdesc[300]{Machinery Fault Detection}
% \ccsdesc{Anomaly Scoring}

% \keywords{Adversarial Learning, Anomaly Detection, Machinery Fault Detection, Reinforcement Learning, Predictive Maintenance}

% \received{9 February 2026}
% \received[revised]{12 March 2026}
% \received[accepted]{5 May 2026}

\maketitle

\section{Introduction} \label{chap4:intor}
Machinery fault detection (MFD) is essential for industrial reliability, yet acquiring labelled fault data for every fault type, severity, and component remains prohibitively expensive. Supervised learning dominates the field ($\approx$81\% of studies, vs.\ just 1\% for RL~\cite{neupane2024machineryDS}) but inherits this labelling bottleneck. Semi-supervised and unsupervised methods partially mitigate it~\cite{ramirez2023semiSupervisedReview, neupane2024MFDReview, 9552620unsupervisedTransferReview, nian2020review} but struggle with robustness under variable operating conditions. Reinforcement learning (RL) is a promising alternative through interactive, sequential decision-making, yet current applications neglect the temporal structure of fault progression~\cite{neupane2024machineryDS}.

Existing ``RL-based'' MFD methods~\cite{ding2019intelligent, qian2022development, li2025novel, li2025convolutional} reduce fault detection to a static \textit{contextual-bandit} (CB) trial: each sensor segment is treated as an independent state, the agent issues a one-shot label, the discount factor $\gamma$ is effectively zero, and episodes terminate after a single action. The sequential premise of RL therefore collapses into supervised classification. Moving beyond this requires multi-step interactions ($\gamma>0$) and rewards reflecting long-term outcomes, but designing such rewards is difficult: fault progression is subtle, system dynamics are only partially understood, and most industrial datasets are prerecorded, making online RL infeasible. Even existing offline RL approaches still rely on manual reward design.

These challenges highlight the potential of inverse reinforcement learning (IRL) \cite{ng2000algorithms}, which bypasses manual reward specification by learning directly from healthy operational trajectories offline. Despite its ability to capture normal machinery behaviour, IRL remains unexplored in machinery fault detection (MFD). To address this gap, we introduce adversarial inverse reinforcement learning (AIRL) \cite{fu2018airl} for MFD, marking its first application in this domain. AIRL operates fully offline, learning a reward model of normal sequential patterns to detect anomalies as deviations from these dynamics. This approach retains RL's strengths while providing a robust, practical solution. Extensive experiments across multiple run-to-failure datasets demonstrate that our framework achieves earlier fault detection and more stable anomaly scoring than conventional one-class and reconstruction-based methods.
\paragraph{Applied contributions.} A two-page extended abstract of this formulation appeared at AAMAS 2026~\cite{neupanelearning} with preliminary HUMS2023 results. This full paper substantially extends that prior work into a comprehensive prognostic system. Our specific contributions are:
(1) a label-free detection pipeline applying the state-only imitation learning (SOIL) approach~\cite{torabi2018generative} to offline industrial sensor sequences; 
(2) validated generalization across three distinct real-world run-to-failure datasets (HUMS2023, IMS, XJTU-SY); 
(3) an extended baseline comparison that demonstrates why AUROC is structurally inapplicable, alongside a novel post-detection consistency (PDC) analysis of \textit{saturation failure}; 
(4) an evaluation of seven automatic thresholding strategies with practical deployment recommendations; 
(5) a computational cost analysis validating AIRL's compact architecture for edge deployment; and 
(6) an ablation study isolating the role of the discount factor in sequential modeling.

\section{Background and Related Works} \label{sec:litrev}
\subsection{Machinery Fault Detection Problem}
Industrial machinery often operates under harsh conditions, resulting in gradual faults—such as wear, cracks, and broken teeth—that vary in type, severity, and the components they affect \cite{nunes2023challenges}. As shown by Neupane et al. \cite{neupane2024MFDReview}, most publicly available datasets reflect this variety, yet are strictly offline and designed for classification tasks with predefined labels. However, in real-world settings, healthy data is abundant, whereas acquiring and labeling faulty data for every possible failure mode is a significant challenge. Consequently, treating MFD as an anomaly detection task, training on abundant normal data to flag deviations at test time, is a convincing and practical approach. While the current research trend remains heavily skewed towards fault classification under supervised learning, robust fault detection remains the foundational necessity in industrial applications.

\subsection{Adversarial Inverse Reinforcement Learning}
Reinforcement learning models sequential decision-making via Markov decision processes (MDPs), optimizing policies to maximize cumulative reward~\cite{sutton2020reinforcement}. However, designing reward functions that precisely encode desired behaviors is challenging, particularly for complex tasks like MFD. IRL addresses this by inferring the underlying reward directly from expert demonstrations~\cite{deshpande2025advances}. While traditional IRL struggles with high-dimensional environments, AIRL~\cite{fu2018airl} scales effectively by employing a generative adversarial framework. AIRL trains a discriminator to differentiate between expert (healthy) and policy-generated trajectories, simultaneously updating a policy to imitate the expert behavior. Importantly, AIRL learns a robust reward function that is disentangled from the environment's specific dynamics, making it highly suitable for anomaly detection where explicit reward formulation is impractical. The discriminator's output is formulated as~\cite{fu2018airl}:

\begin{equation}
D(s, a, s') =
\frac{\exp\!\left(r(s, a) + \gamma\,V(s') - V(s)\right)}
     {\exp\!\left(r(s, a) + \gamma\,V(s') - V(s)\right)
      + \pi(a \mid s)},
\label{eq:airl_disc}
\end{equation}
where $(s, a, s')$ is the transition tuple consisting of the current state $s$, action $a$, and next state $s'$, $r(s, a)$ is the learned reward, $V(s)$ the state-value function, $\pi(a \mid s)$ the stochastic policy, and $\gamma$ the discount factor. This architecture ensures the recovered reward remains invariant to dynamic shifts, providing a stable basis for evaluating operational health.

\subsection{RL {vs.} Contextual-Bandit Practice in MFD}
\label{subsec:relatedWork}
The application of RL to rotating machinery fault detection has gained popularity since 2019, as summarized in Table~\ref{tab:rl-mfd-survey}. However, a consistent pattern emerges: despite the variety of agents and reported episode lengths, these studies formulate the problem as a single-step classification task rather than a true MDP. Specifically:
\begin{itemize}
\item Datasets are typically pre-shuffled, treating each sensor signal segment as an \textit{iid} (independent and identically distributed) context. The agent predicts a single label, and the episode immediately terminates.
\item Because transitions $(s,a,s')$ are ignored, the discount factor $\gamma$ (even if nonzero) plays no mathematical role in credit assignment.\end{itemize}
Formally, these setups are \textit{single-step contextual bandits (CBs)}. Even when studies report episode lengths of several hundred steps (Table~\ref{tab:rl-mfd-survey}), each step remains an independent sample with no dependency on previous actions. The Bellman equation collapses into a supervised classifier optimized via a reward-based loss. While this approach is valuable for static labelling, using RL terminology misrepresents the methodological novelty. More importantly, it fails to exploit RL's core strength: modelling sequential system evolution for fault \textit{progression} and \textit{early-warning} objectives.
\begin{table}[!htbp] % CHANGED [t] to [!htbp]
\centering
\caption{RL formulations in recent MFD research.}
\label{tab:rl-mfd-survey}
\resizebox{\columnwidth}{!}{%
\begin{tabular}{@{}lccccc@{}}
\toprule
\textbf{Ref/Year} & \textbf{State} & \textbf{Action} & $\boldsymbol{\gamma}$ & \textbf{Ep. Len.} & \textbf{Agent}\\
\midrule
\cite{ding2019intelligent}/19 & Random & Label & 0 & 1 & SAE \\
\cite{li2021deep}/21 & Random & Label & - & 1 & CapNet \\
\cite{qian2022development}/22 & Random & Label & 0 & 1 & CNN-GRU \\
\cite{yang2023new}/23 & Random & Label & 0.1* & 500\textsuperscript{†} & ConvResAE \\
\cite{xiao2024research}/24 & Random & Label & 0.1* & 504\textsuperscript{†} & 1D-CNN \\
\cite{li2025novel}/25 & Random & Label & 0.99* & 1 & Eq. DQN \\
\cite{li2025convolutional}/25 & Random & Label & 0.99* & 1 & ConvTrans \\
\cite{saied2025development}/25 & Random & Label & 0.1* & 1 & Duel-DQN \\
\bottomrule
\end{tabular}%
}
{\scriptsize \raggedright \textbf{Note:} * $\gamma$ not used for credit assignment. \textsuperscript{†} Samples pre-shuffled. \par}
\vspace{-4 mm}
\end{table}

\section{AIRL Framework for MFD} \label{sec:airl_framework}
Building on the problem formulation and related work discussed earlier, this section presents the proposed adversarial inverse reinforcement learning framework for machinery fault detection.
The framework uses preprocessed vibration sequences to autonomously learn reward dynamics that characterize normal operational behavior and deviations from it.
It consists of three main components: (i) transition construction from vibration signals, (ii) neural architectures for reward, value, and policy estimation, and (iii) adversarial training and threshold-based anomaly scoring.

\subsection{Transition Construction for AIRL}
Let $X = \{x_1, x_2, \dots, x_L\}$ denote a vibration record of length $L$ samples, and let $\tilde{x}_t$ denote the $z$-normalized version of $x_t$ using statistics computed solely from the healthy training partition (to prevent test-train leakage). Records are sorted chronologically and split into a healthy expert set $\mathcal{D}_{\text{exp}}$ used for training and a held-out test set $\mathcal{D}_{\text{test}}$ containing the degradation phase. Dataset-specific partitions, record lengths, and acquisition details are reported in Section~\ref{sec:experimental}.

To construct the AIRL input, state-action-next-state tuples are generated from each normalized sequence:
\begin{equation}
    s_{t} = \tilde{x}_{t}, \quad a_{t} = \tilde{x}_{t+1}, \quad s_{t+1} = \tilde{x}_{t+1}.
\end{equation}
Since the data is offline and purely observational, no recorded control actions are available. We adopt the SOIL approach~\cite{torabi2018generative} and treat the next observation $\tilde{x}_{t+1}$ as a proxy action, effectively learning an inverse-dynamics model that scores the plausibility of state transitions $(s_t, s_{t+1})$ rather than maximizing a static classification reward. This aligns with state-only imitation and inverse-RL methods such as GAIfO~\cite{torabi2019adversarial}, Diffusion-IfO~\cite{NEURIPS2024_f7faa46b}, and OPOLO~\cite{zhu2020off}, which optimize a discriminator over state transitions rather than explicit control actions. The construction enables AIRL to model the temporal evolution of healthy and faulty sequences without introducing artificial control variables.

\subsection{Model Architecture}
\label{subsec:airl_architecture}
The proposed AIRL framework comprises three neural components: a reward network $r(s,a)$, a value network $V(s)$, and a stochastic policy $\pi(a|s)$. Note that while we denote the input as $a$ to maintain standard RL notation, $a$ corresponds to the next state $\tilde{x}_{t+1}$. Each network is a two-layer MLP of the form $f(x) = W_2\,\phi(W_1 x + b_1) + b_2$, where $\phi(\cdot)$ is the ReLU activation, the hidden layer has 64 units, and $b_1, b_2$ are learnable biases. The networks differ only in their input dimension and output role:
\begin{itemize}
    \item \textbf{Reward network ($r$):} Maps the concatenated state-action vector $[s;a]\in\mathbb{R}^2$ to a scalar reward distinguishing healthy (expert) from faulty (policy-generated) transitions.
    \item \textbf{Value network ($V$):} Takes scalar state $s\in\mathbb{R}$ and outputs the expected cumulative reward, providing a potential-based shaping term for reward generalization.
    \item \textbf{Policy network ($\pi$):} Takes scalar state $s\in\mathbb{R}$ and outputs the mean $\mu(s)$ of a Gaussian policy $\pi(a|s) = \mathcal{N}(\mu(s), \sigma^2)$; the state-independent log-standard-deviation $\log\sigma$ is learned separately. This stochasticity encourages exploration during roll-outs and mitigates overfitting to logged demonstrations.
\end{itemize}
The total parameter count is 644 (reward: 257, value: 193, policy: 194).

\subsection{Training Procedure}
\label{subsec:airl_training}
Each AIRL training epoch consists of two phases: (i) discriminator learning and (ii) policy improvement.
During discriminator learning, mini-batches of 128 expert and policy-generated transitions are drawn, and the reward and value networks are jointly updated by minimizing the standard AIRL cross-entropy loss:
% \begin{equation}
% \resizebox{\columnwidth}{!}{$
% \mathcal{L}_{\text{disc}} = - \biggl[ \log \sigma\bigl(r_{\phi}(s,a) + \gamma V_{\psi}(s') - V_{\psi}(s)\bigr) + \log \Bigl(1 - \sigma\bigl(r_{\phi}(s,a_{\pi}) + \gamma V_{\psi}(s'_{\pi}) - V_{\psi}(s_{\pi})\bigr)\Bigr) \biggr]
% $}
% \end{equation}

\begin{align}
\mathcal{L}_{\text{disc}} =\;
& - \biggl[
    \log \sigma\bigl(r_{\phi}(s,a) + \gamma V_{\psi}(s') - V_{\psi}(s)\bigr) \nonumber \\
& \qquad +\; \log \Bigl(1 - \sigma\bigl(r_{\phi}(s,a_{\pi}) + \gamma V_{\psi}(s'_{\pi}) - V_{\psi}(s_{\pi})\bigr)\Bigr)
  \biggr]
\end{align}
where $\sigma(\cdot)$ is the logistic function and $\gamma=0.99$.

During policy improvement, $\pi$ is rolled out in the learned reward landscape to estimate the REINFORCE gradient:
\begin{equation}
\nabla\,\mathbb{E}_{\pi}\![r(s,a)] = \mathbb{E}_{\pi}[\nabla\log\pi(a|s)\,G_t],
\end{equation}
using 50-step returns $G_t$, alongside an entropy bonus $\beta=0.01$ to encourage exploration. All networks are optimized with Adam ($\alpha=10^{-3}$) for $N_{\text{epochs}}=300$ with 50 policy episodes per epoch.

\subsection{Anomaly Scoring and Thresholding}
\label{subsec:anomThresholds}
Within AIRL, the discriminator assigns each transition $(s,a,s')$ a confidence $D_\phi(s,a,s')$, which we convert to an anomaly score:
\begin{equation}
\alpha(s,a,s') = 1 - D_\phi(s,a,s'),
\end{equation}
where higher values indicate greater deviation from expert behavior. Let $\{\alpha_i\}_{i=1}^{N}$ denote scores computed on \textit{healthy} training transitions. We construct data-driven thresholds $\tau$ from these healthy scores and flag a test sample as anomalous when $\alpha>\tau$ (upper-tail rule).

Seven strategy families are evaluated: (1)~mean+$k\sigma$ ($k\in\{2,3\}$) \cite{neupane2024comparative}; (2)~median+3MAD \cite{romo2024median}; (3)~max-based variants (max, max$-$SE, max$-$2SE) \cite{yang2025agent}; (4)~percentile $p_{99}$ \cite{umsonst2022finite}; (5)~Otsu's method \cite{otsu1979threshold}; (6)~$k$-means midpoint \cite{macqueen1967kmeans}; (7)~GPD tail fit (top 5\% of $\{\alpha_i\}$) \cite{pickands1975statistical}. All thresholds are computed once from healthy training scores and fixed at test time.

\section{Experimental Section} \label{sec:experimental}
\subsection{Datasets and Preprocessing} \label{subsec:preprocessingHUMS}

\begin{figure}
\centering
\includegraphics[width=0.5\textwidth]{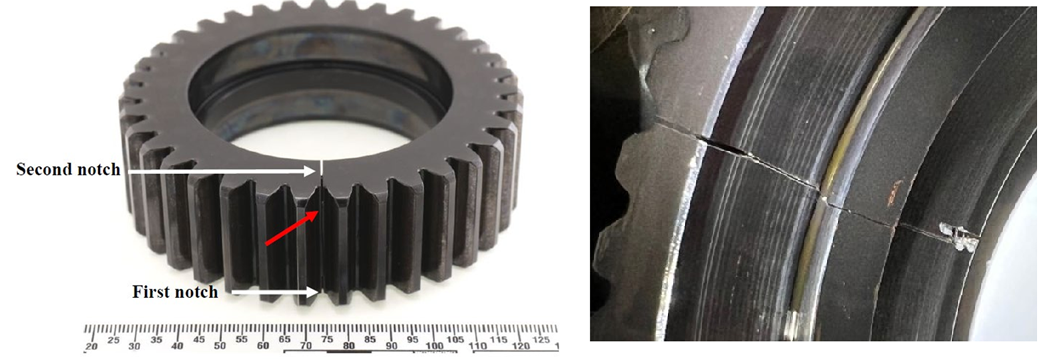}
\caption{Cracked planet gear: crack path indicated by red arrow (left), and full crack viewed from the inner bore of the planet gear’s bearing raceway (right) \cite{wang2023helicopter}.}
\label{fig:humsFaultCrack}
\end{figure}

We evaluate the proposed framework on three real-world run-to-failure datasets encompassing distinct machinery types, operating conditions, and fault modes. All methods use healthy-only training partitions and mixed (healthy + faulty) test partitions, as detailed in Table~\ref{tab:train_test_splits}.

\textbf{HUMS2023}~\cite{wang2023helicopter}: This dataset captures the progressive degradation of a full-scale helicopter gearbox under authentic load conditions. Fatigue cracks were initiated via EDM notches and propagated under 125\% torque (see Figure~\ref{fig:humsFaultCrack}. Hunting Tooth Synchronous Averaging (H-SSA) was utilized to align the vibration patterns with gear-meshing cycles~\cite{sawalhi2024helicopter}. The ground truth degradation onset is verified by the challenge committee as Day~24 (sample~\#264).

\textbf{IMS Test~1}~\cite{sacerdoti2023comparison}: This is a widely used bearing dataset recorded at 20\,kHz across 8 channels, from which Channel~1 is utilized for our evaluation. The ground truth degradation onset is established in the literature as mid-November~\cite{sacerdoti2023comparison,berghout2021leveraging}, which eventually culminates in catastrophic failure on Nov~25.

\textbf{XJTU-SY}~\cite{wang2020xjtusy}: This dataset features accelerated bearing degradation under heavy operational loads (Bearing 3\_1; 40\,kHz, 10\,kN). The ground truth degradation onset is formally defined at $\approx$500 min~\cite{mao2024harmony}.

\begin{table}[htbp] % Changed [t] to [htbp] for better automated placement
\small
\centering
\renewcommand{\arraystretch}{1.2}
\caption{Training and testing splits.}
\label{tab:train_test_splits}
\begin{tabular}{l|c|c}
\hline
Dataset & Training (healthy) & Testing (mixed) \\
\hline
HUMS & Day~17--20; Total: 282 & Day~21--27; Total: 526 \\
IMS   & Oct~1--31; Total: 1,840 & Nov~1--25; Total: 8,940 \\
XJTU--SY & First 2,000 ($\sim$250 min) & Remaining~18,295 \\
\hline
\end{tabular}
\end{table}

To maintain consistent temporal resolution and input dimensionality across the entire framework, all raw vibration sequences are $z$-normalized and segmented into uniform observation windows (4095 points for HUMS H-SSA samples, and 4096-point segments for IMS and XJTU-SY, yielding 5 segments per IMS sample and 8 per XJTU-SY sample).

\subsection{Baseline Methods}
For comparison with AIRL, we trained a set of widely used unsupervised and one-class methods using only healthy data and evaluated them on both healthy and faulty segments. These baselines encompass a range of approaches, from traditional machine learning to deep learning models designed to capture temporal dependencies.

\subsubsection{Unsupervised and One-Class Baselines}
All baselines are trained on healthy data only and scored with the same thresholding rules as AIRL.
\begin{itemize}
    \item \textbf{IF} (Isolation Forest) \cite{liu2008isolation} and \textbf{OCSVM} (One-Class Support Vector Machine) \cite{scholkopf2001estimating}: classical one-class methods. {IF: 400 estimators, max\_samples=256; OCSVM: RBF kernel, $\nu=0.05$.}
    \item \textbf{AE} (Autoencoder) \cite{zhou2017anomaly} and \textbf{VAE} (Variational Autoencoder) \cite{kingma2013auto}: reconstruction-error anomaly scorers (hidden dims \{1024,256,64\}, latent 64). {($\sim$8.95M / $\sim$8.97M params)}
    \item \textbf{LSTM-AE} and \textbf{LSTM-VAE} \cite{malhotra2016lstm,chung2015vrnn}: Long Short-Term Memory (LSTM) recurrent variants of AE/VAE for sequence modeling. {(LSTM hidden=128, latent=64; $\sim$167K / $\sim$184K params)}
    \item \textbf{TCAE} (Temporal Convolutional Autoencoder) \cite{bai2018empirical}: 1D convolutional AE with strided convolutions (spatial $4096\!\to\!64$ via three stride-4 Conv1d layers, kernel 9, channels 1$\to$16$\to$32$\to$8). {($\sim$14K params)}
    \item \textbf{CTCAE} (Contrastive TCAE) \cite{chen2020simple}: TCAE + SimCLR contrastive loss on noise-augmented views. {(wider bottleneck: channels 1$\to$16$\to$32$\to$64, $\sim$47K backbone params)}
    \item {\textbf{USAD} (Unsupervised Anomaly Detection) \cite{audibert2020usad}: dual-decoder adversarial AE; anomaly score combines both decoders' outputs ($\sim$13.4M params).}
    \item {\textbf{ATF} (Anomaly Transformer) \cite{xu2022anomaly}: uses association-discrepancy between a Gaussian prior and learned softmax attention; ConvStem tokenizes inputs into 91 tokens for HUMS (4095 pts) and 92 tokens for IMS/XJTU-SY (4096 pts) (2 blocks, 64 dim, 4 heads, $\sim$146K params).}
\end{itemize}

\subsubsection{Contextual-Bandit Baseline}
\label{subsec:contexualBandit}
We reproduce the LiteDPER--CTQN agent~\cite{li2025convolutional} (reimplemented in PyTorch) as a representative of the common RL-as-classifier pattern in MFD. It trains on balanced Normal/Faulty splits (HUMS: Days~19--20 vs.\ 26--27; IMS: 1370 Oct segments vs.\ last 1370 Nov segments; XJTU-SY: first 2000 vs.\ last 2000 segments) and is tested on the remaining unlabeled files.

\subsection{Evaluation Criteria}
\label{subsec:evaluation_criteria_auroc}
Our primary objective is evaluating \textit{earliest detection}, i.e., how soon a model reliably signals anomalies within the incipient degradation window. Because early warnings are operationally useless alongside continuous false positives, we also assess the pre-onset \textit{False Alarm Rate} (FAR) and \textit{post-detection consistency} to ensure the stability of valid alerts.

\textbf{Why AUROC is inapplicable:} Run-to-failure data lacks per-sample binary labels, and retrospective fault onsets are inherently uncertain (e.g., the HUMS2023 onset spans Days~22--24). Consequently, AUROC computed against a rigid boundary is fragile and indifferent to detection earliness. Following standard practices~\cite{lei2018machinery,li2018remaining}, we instead evaluate reliability via \textit{timeliness}, \textit{FAR}, and \textit{PDC}, explicitly flagging \textit{saturation failure} (achieving high PDC merely through persistent, indiscriminate alerting).

\subsection{Experimental Setup}
All models use the splits in Table~\ref{tab:train_test_splits}, trained over ten random seeds (scores averaged). AIRL uses the hyperparameters in Section~\ref{subsec:airl_training}. AE-family baselines use symmetric encoder–decoders with hidden dims \{1024,256,64\}, latent 64. IF uses 400 estimators; OCSVM uses an RBF kernel. All methods share the same thresholding rules (Section~\ref{subsec:anomThresholds}).

\section{Evaluation} \label{sec:analysis}
\subsection{Earliest Fault Detection} \label{subsec:earliestDetection}

\begin{figure}
\centering
\includegraphics[width=0.5\textwidth]{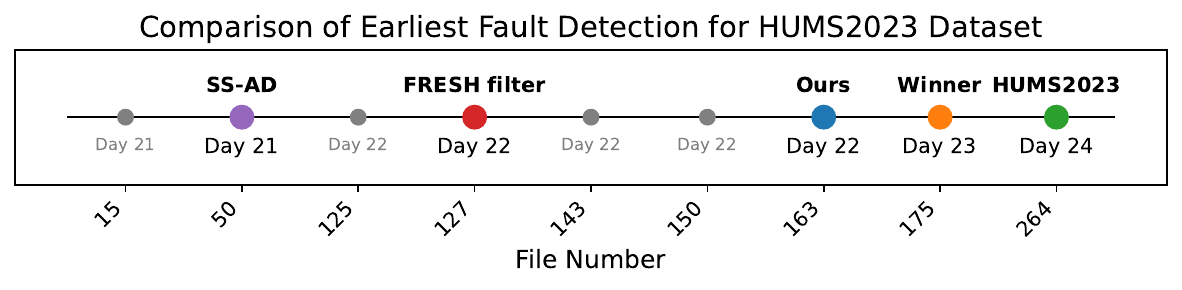}
\caption{Comparison of Earliest Fault Detection for HUMS2023 Dataset}
\label{fig:hums_result_AIRL}
\end{figure}

\begin{table}[htbp]
\centering
\caption{Earliest fault detection across datasets (max$-$2SE threshold).}
\label{tab:earliest_detection}
\resizebox{\columnwidth}{!}{%
\begin{tabular}{@{}lccc@{}}
\toprule
\textbf{Model} & \textbf{HUMS} & \textbf{IMS} & \textbf{XJTU-SY} \\
\cmidrule(lr){2-4}
\makecell[l]{{{ GT}}} & Day~24~\#264 & mid-Nov & $\sim$500 min\\
\midrule
IF          & Day~21~\#9      & 2003-11-08~11:41:44 & 279 \\
OCSVM       & Day~21~\#1      & 2003-11-01~00:41:44 & 264 \\
AE          & Day~21~\#1      & 2003-11-01~05:31:44 & 309 \\
VAE         & Day~21~\#37     & 2003-11-22~17:36:56$^\dagger$ & 332 \\
LSTM-AE     & Day~21~\#36     & 2003-11-22~17:36:56$^\dagger$ & 332 \\
LSTM-VAE    & Day~21~\#37     & 2003-11-22~17:36:56$^\dagger$ & 332 \\
TCAE        & Day~21~\#49     & 2003-11-03~11:51:44 & 265 \\
CTCAE       & No~Fault  & No~Fault & No~Fault \\
USAD        & Day~21~\#9 & 2003-11-01~11:11:44 & 264 \\
ATF         & Day~21~\#37 & 2003-11-14~11:12:17 & 310 \\
CTQN (CB)   & Day~21~\#1$^{\ddagger}$ & 2003-11-01~01:21:44 & 455 \\
AIRL ($\gamma=0$) & { Day~22~\#163} & { 2003-11-09~01:01:44} & { 277} \\
AIRL (Ours) & {Day~22~\#163} & {{2003-11-17~12:22:30}} & {459} \\
\bottomrule
\end{tabular}%
}

{\scriptsize \raggedright 
\textit{GT:} Provided Ground Truth. \textit{AIRL ($\gamma=0$):} Ablation results (Sec \ref{sec:appendix}). $^\dagger$\textit{Post-GT:} Misses onset window under strict threshold. $^{\ddagger}$\textit{CTQN:} Binary classification collapse (all-Faulty constant score $\approx$0.546).\par}
\vspace{-1.5em} % Aggressively pulls up the following text
\end{table}

Table~\ref{tab:earliest_detection} summarizes the earliest anomaly detections across the three datasets. On HUMS, nearly every baseline triggered on Day~21 (the first day of the test partition, which contains 89 files indexed \#1--\#89; baseline detections clustered in the early portion of this range, between \#1 and \#49), well before the committee-verified onset of Day~24 \#264, indicating premature false alarms. {CTQN exhibited mode collapse: all 416 HUMS test files (Days~21--25) flagged faulty with a near-constant score $\approx$0.546, while CTCAE failed in the opposite direction, producing no detections at all. AIRL detected at Day~22 \#163, late enough to clear the false-alarm zone, early enough to precede committee ground truth. Figure~\ref{fig:hums_result_AIRL} illustrates this against prior HUMS2023 work: AIRL falls between FRESH filter~\cite{vaerenberg2025detecting} (Day~22 \#127) and the official challenge winner~\cite{peeters2024fatigue} (Day~23 \#175), aligning with the committee-verified onset.

On IMS, evaluated against the mid-November onset~\cite{berghout2021leveraging,sacerdoti2023comparison} (catastrophic failure Nov~25), baselines split into three failure modes. OCSVM, AE, USAD, and CTQN triggered Nov~1, with TCAE on Nov~3, indicating pre-GT false alarms days before any physical degradation. IF and ATF (Nov~8 and Nov~14) fall earlier than the established window. Conversely, VAE, LSTM-AE, and LSTM-VAE first flag only on Nov~22, missing the prognostic window entirely. AIRL alone triggered on \textit{Nov~17}, precisely within the mid-November onset. The ablated AIRL ($\gamma=0$) flagged on Nov~9, behaving like the statistical baselines and confirming that the discount factor is what separates genuine degradation from operational noise.

On XJTU-SY (healthy window ends at $\approx$500~min~\cite{mao2020robust}), every baseline except CTQN triggered between Files~264--332, more than 150~min before the onset boundary and well within the established healthy range; these are clear pre-onset false alarms. CTCAE failed to register any deviation. Two methods produced detections close to the onset boundary: CTQN at File~455 and AIRL at File~459. Given that the $\sim$500-min reference is itself an approximate onset annotation rather than a hard cutoff, both detections fall within a 45-min precursor window of the verified degradation transition, and are best interpreted as operationally reasonable early warnings rather than false alarms.} On this dataset alone, CTQN's bandit formulation produced a comparably reasonable detection point; however, its performance on HUMS (mode collapse) and IMS (false alarm Nov~1) shows this is not a stable property of the formulation. AIRL achieves comparable near-onset detection on XJTU-SY \textit{and} stable behaviour on the other two datasets, which is the broader claim the bandit baseline cannot match.

No single method dominates every individual benchmark, e.g., ATF achieves the lowest FAR on IMS under Otsu (Section~\ref{subsec:consistency_stability}). What AIRL alone achieves is cross-dataset reliability: across all three datasets, AIRL avoids both failure modes the baselines exhibit: the oversensitivity of reconstruction and one-class methods (premature false alarms, or onset-missing as VAE and LSTM variants do on IMS), and the instability of static binary classifiers and contrastive methods (mode collapse, saturation, or total detection failure). This is exactly what the contextual-bandit analysis in Section~\ref{subsec:relatedWork} predicted: without state transitions or temporal credit assignment, an agent cannot stabilize on a meaningful decision boundary. AIRL's combination of sequential reasoning and adversarial reward learning is what delivers timely, trustworthy early warnings consistently across gearbox and bearing fault types.

% Across all three datasets, AIRL avoids both failure modes the baselines exhibit: the oversensitivity of reconstruction and one-class methods (premature false alarms, or onset-missing as VAE and LSTM variants do on IMS), and the instability of static binary classifiers and contrastive methods (mode collapse, saturation, or total detection failure). This is exactly what the contextual-bandit analysis in Section~\ref{subsec:relatedWork} predicted: without state transitions or temporal credit assignment, an agent cannot stabilize on a meaningful decision boundary. AIRL's combination of sequential reasoning and adversarial reward learning is what delivers timely, trustworthy early warnings consistently across gearbox and bearing fault types.

\subsection{Post-Detection Consistency}
\label{subsec:consistency_stability}

\begin{table}[h]
% \vspace{-1 em}
\centering
\caption{Pre-GT FAR and PDC after Ground Truth (Otsu).}
\label{tab:pdc_post_gt}
\resizebox{\columnwidth}{!}{%
\begin{tabular}{@{}lcccccc@{}}
\toprule
\multirow{2}{*}{\textbf{Method}}
  & \multicolumn{2}{c}{\textbf{HUMS}}
  & \multicolumn{2}{c}{\textbf{IMS}}
  & \multicolumn{2}{c}{\textbf{XJTU-SY}} \\
\cmidrule(lr){2-3}\cmidrule(lr){4-5}\cmidrule(lr){6-7}
 & {\textbf{FAR\%}} & \textbf{PDC\%}
 & {\textbf{FAR\%}} & \textbf{PDC\%}
 & {\textbf{FAR\%}} & \textbf{PDC\%} \\
\midrule
IF        & {[95.1]}  & {[99.62]}  & {57.2}   & {59.70}  & {50.3}    & 50.52 \\
OCSVM     & {[99.2]}  & {[100.0]}  & {60.3}   & {61.86}  & {58.9}    & 59.18 \\
AE        & {[100.0]} & {[100.0]}  & {82.6}   & {90.73}  & {[97.8]}  & {[98.55]} \\
VAE       & {89.1}    & {99.86}   & {50.0}   & {49.91}  & {46.8}    & 46.85 \\
LSTM-AE   & {[93.2]}  & {[99.24]}  & {50.2}   & {50.18}  & {46.5}    & 46.58 \\
LSTM-VAE  & {[91.6]}  & {[99.24]}  & {50.2}   & {50.12}  & {46.5}    & 46.58 \\
TCAE      & {41.8}    & 68.06      & {88.3}   & {92.68}  & {86.3}    & 87.62 \\
CTCAE     & {0.0}     & 0.00       & {0.0}    & 0.00     & {0.0}     & 0.00 \\
CTQN (CB) & {[100.0]} & {[100.0]} & {52.7} & {90.0} & {63.0} & {100.0} \\
USAD      & {[99.8]}  & {[99.8]}   & {70.1}   & 70.9     & {81.4}    & 81.5 \\
ATF       & {[94.9]}  & {[96.32]}  & {11.6}   & {94.3}    & {[100.0]} & {[81.79]} \\
{AIRL (Ours)} & {{42.2}} & {79.09} & {{88.1}} & {90.15} & {{68.4}} & {68.38} \\

\bottomrule
\end{tabular}%
}
% \vspace{1mm}
{\footnotesize \raggedright 
\textit{[ ]} denotes saturation failure (FAR $>$ 90\%); high PDC reflects false-alarm persistence, not genuine discrimination.\par}
\end{table}

Beyond detection timing, a practical prognostic system must remain
stable throughout the degradation period. \textit{Post-detection
consistency} measures the proportion of files correctly
flagged as anomalous after the confirmed ground-truth
onset~\cite{lei2018machinery,li2018remaining}; a high PDC indicates
that the model tracks fault progression steadily without frequent
dropouts. Critically, PDC must always be interpreted alongside
detection timing: a method that triggers prematurely and never
stops alarming will accumulate a near-perfect PDC through
false-alarm persistence, not genuine discrimination. We term this
\textit{saturation failure}, and it renders the PDC score
functionally meaningless for that method.

\paragraph{Two-threshold evaluation framework.}
Table~\ref{tab:earliest_detection} (earliest detection) uses the
\textit{max}$-$2SE threshold, which is calibrated from the maximum
healthy training score and is one of the max-based strategies producing earliest within-window detection across all three datasets in Table~\ref{tab:threshold_sensitivity}; it answers
``\textit{when} does the method first alarm under
strict, max-based calibration?'' Table~\ref{tab:pdc_post_gt} PDC
uses Otsu's adaptive threshold, which partitions test scores
globally to maximize between-class variance; it answers
``\textit{how reliably} does the method track fault progression
after onset?''

\textbf{HUMS2023 Gearbox:} All reconstruction-based baselines achieve $>$99\% PDC on HUMS under the Otsu threshold,
but only because they trigger prematurely on Day~21 and flag
virtually every subsequent file, collapsing to an
``always-anomalous'' state. USAD exhibits
the same pattern across all three datasets. ATF saturates on HUMS (FAR=95\%). CTCAE scores 0\% PDC on HUMS, indicating total detection failure. CTQN scores 100\% PDC on HUMS but with 100\% FAR (all files flagged Faulty from Day~21~\#1), confirming saturation failure rather than genuine detection; its high PDC is as meaningless as any other saturated method. TCAE achieves 68.06\% PDC on HUMS, with its bearing-dataset numbers similarly inflated by pre-GT false-alarm persistence (Table~\ref{tab:pdc_post_gt}).

\textbf{IMS Bearings:} On IMS, AIRL achieves PDC of 90.15\% with a FAR of 88\%. ATF achieves a low FAR (12\%) on IMS but collapses to zero post-GT detections under strict thresholds, confirming its IMS score range is inadequately spread for conservative calibration. TCAE and IF achieve competitive PDC (92.68\%, 59.70\%) but at the cost of high pre-GT FAR (88\%, 57\% respectively), confirming that their IMS performance is inflated by persistent flagging begun well before degradation onset. CTQN achieves PDC of 90.0\% with a FAR of 52.7\% on IMS, with detection from Nov~1 (01:21:44), well before the mid-November onset, indicating its IMS PDC is partially inflated by early false-alarm persistence beginning from the first test day.

\textbf{XJTU-SY Bearings:} On XJTU-SY, ATF completely saturates (FAR=100\%) while AIRL maintains a FAR of 68\% with a PDC of 68.38\%. The compressed score range of this bearing dataset makes Otsu partitioning less precise for all methods; nonetheless, AIRL is the only method achieving consistent non-saturated behavior across this dataset as well. CTQN achieves PDC of 100\% with FAR of 63.0\% on XJTU-SY, with first detection at minute~455 (pre-GT), suggesting competitive post-fault tracking, but this is evaluated on a 16{,}295-file subset (the last 2{,}000 test files are withheld as training data), so the post-GT window differs from other methods.

AIRL (Ours) is the only method achieving meaningful, non-saturated PDC across all three datasets: 79.09\% (HUMS), 90.15\% (IMS), and 68.38\% (XJTU-SY). On HUMS, AIRL significantly outperforms every non-saturated baseline. IMS is AIRL's weakest dataset: the pre-GT FAR of 88\% reflects that gradual bearing degradation produces a compressed AIRL score range, where Otsu's adaptive split places the decision boundary close to the healthy distribution. ATF achieves a lower IMS FAR (12\%) under Otsu but saturates on HUMS (FAR=94.9\%) and XJTU-SY (FAR=100\%), and collapses to zero post-GT detections under strict thresholds (Table~\ref{tab:earliest_detection}); its IMS success is dataset-specific, not a general property. TCAE matches AIRL's IMS FAR (88\%) but is itself saturated on HUMS and XJTU-SY. AIRL is the only method whose FAR stays below the 90\% saturation ceiling on every dataset, demonstrating that the temporal reward signal generalises across distinct fault types where reconstruction and attention-based baselines do not. High reconstruction capacity does not prevent saturation-or-collapse; temporal reward modeling does.
\vspace{-1 em}
\vspace{-1mm}

\subsection{Threshold Sensitivity Analysis}
\label{subsec:threshold_sensitivity}
{Threshold choice is a practical deployment concern. Table~\ref{tab:threshold_sensitivity} compares the earliest AIRL detection under all threshold strategy families described in Section~\ref{subsec:anomThresholds}.}

On HUMS, five independent strategies converge to the same detection point (Day~22, \#163): mean+3$\sigma$, all three max-based variants, and GPD tail ($p_{99}$). Percentile $p_{99}$ is also within the onset window (Day~22~\#100). This cross-strategy consistency confirms that the AIRL reward signal produces a clean, threshold-invariant separation on gearbox data. Loose strategies (Otsu, $k$-means) trigger on Day~21~\#2.

On IMS, the threshold landscape is sharply polarised. All loose and moderate strategies (mean+$k\sigma$, median+3MAD, percentile, Otsu, $k$-means) trigger false alarms from Nov~1, well before the mid-November degradation window, while GPD tail ($p_{99}$) triggers on Nov~8. In contrast, all three max-based strategies (Max, Max$-$SE, Max$-$2SE) converge to exactly one anomalous file across the entire 8,940-file test set: Nov~17 12:22:30, precisely within the mid-November onset window. This binary behaviour, either saturation or single-point detection, reflects the compressed AIRL score range on IMS, where the Nov~17 file is the sole test sample that marginally exceeds the healthy training maximum.

On XJTU-SY, the usable strategy space is narrower due to the compressed score range: mean+2$\sigma$, mean+3$\sigma$, median+3MAD, and Max yield no detection because no test file exceeds the corresponding threshold. The reported onset is approximately 500 minutes (end of the healthy window); the precise transition is subtle. Both {Max$-$2SE} (minute~459) and {Max$-$SE} (minute~533) detect within a 75-minute window of this boundary, with Max$-$2SE flagging the earliest pre-onset deviation and Max$-$SE confirming detection just after. For prognostic deployments where early warning matters, Max$-$2SE is preferable; for confirmed-fault detection, Max$-$SE is more conservative. GPD tail ($p_{99}$) is excessively delayed (minute~1748), and Otsu and $k$-means saturate from minute~251.

Across all three datasets, the {max-based family (Max$-$SE and Max$-$2SE)} provides the most reliable detection. Both variants produce identical detections on HUMS (Day~22~\#163) and IMS (Nov~17), and bracket the XJTU-SY onset by under 75 minutes. We recommend {Max$-$2SE} as the default for prognostic early warning (earliest detection within the onset window across all three datasets), and Max$-$SE as the conservative alternative when post-onset confirmation is preferred. Both are self-calibrating from healthy training scores and require no held-out validation set or manual tuning.

\begin{table}[h]
\centering
\caption{Earliest AIRL detection per threshold strategy}
\label{tab:threshold_sensitivity}
\resizebox{\columnwidth}{!}{%
\begin{tabular}{@{}lccc@{}}
\toprule
\textbf{Threshold} & \textbf{HUMS (Day\,\#)} & {\textbf{IMS (date)}} & {\textbf{XJTU-SY (min)}}\\
\midrule
Mean+2$\sigma$    & Day~21~\#88{$\ddagger$}   & {Nov~1$\ddagger$}  & {---} \\
Mean+3$\sigma$    & Day~22~\#163{$\dagger$}   & {Nov~1$\ddagger$}  & {---} \\
Median+3MAD       & Day~21~\#80{$\ddagger$}   & {Nov~1$\ddagger$}  & {---} \\
Max               & Day~22~\#163{$\dagger$}   & {Nov~17$\dagger$}  & {---} \\
Max$-$SE          & Day~22~\#163{$\dagger$}   & {Nov~17$\dagger$}  & {533$\dagger$} \\
Max$-$2SE         & Day~22~\#163{$\dagger$}   & {Nov~17$\dagger$}  & {459$\dagger$} \\
Percentile ($p_{99}$) & Day~22~\#100{$\dagger$} & {Nov~1$\ddagger$} & {267$\ddagger$} \\
Otsu              & Day~21~\#2{$\S$}          & {Nov~1$\S$}        & {251$\S$} \\
$k$-means midpoint & Day~21~\#2{$\S$}         & {Nov~1$\S$}        & {251$\S$} \\
GPD tail ($p_{99}$) & Day~22~\#163{$\dagger$} & {Nov~8$\ddagger$}  & {1748$\dagger$} \\
\midrule
Ground truth onset & Day~24~\#264 & {mid-Nov.}          & {$\sim$500 min} \\
\bottomrule
\end{tabular}%
% }
% \vspace{-1mm}
% {\footnotesize \raggedright 
% $^\dagger$ \textit{Valid Detection:} Within-window / post-GT detection. \\
% $^\ddagger$ \textit{False Alarm:} Pre-GT false alarm. \\
% $^\S$ \textit{Saturation:} Triggered from the first test file. \\
% --- \textit{No Detection:} No test file exceeds the threshold.\par}
% \vspace{-3 mm}
% \end{table}
}
% \vspace{-1mm}
{\footnotesize \raggedright 
$^\dagger$\textit{Valid:} Within-window / post-GT. \quad $^\ddagger$\textit{False Alarm:} Pre-GT. \\
$^\S$\textit{Saturation:} Flags first test file. \quad ---\textit{No Detection:} Threshold not met.\par}
\vspace*{-3mm}
\end{table}

\subsection{Computational Cost and Deployment}
\label{subsec:computational_cost}
Adversarial training is commonly cited as expensive; in practice, AIRL's overhead is modest due to two structural properties. First, its three networks (reward, value, policy) contain only 644 parameters, one to four orders of magnitude fewer than the deep baselines (TCAE $\sim$14K, LSTM-AE $\sim$167K, ATF $\sim$146K, AE $\sim$9.0M, USAD $\sim$13.4M). Second, at inference time AIRL operates on consecutive scalar pairs $(x_t, x_{t+1})$: each full-length file is reduced to a single transition pair before scoring, requiring no matrix multiplication over high-dimensional vectors. All reconstruction-based methods (AE, TCAE, USAD, ATF) must forward-pass the full sequence vector per file. This makes AIRL's per-file inference constant, minimal, and compatible with edge controllers that cannot buffer complete sample windows. For deployment: (i)~collect $\geq$250 healthy files; (ii)~train AIRL offline once on healthy transitions; (iii)~at runtime, score each new acquisition file using the frozen reward and value networks; (iv)~apply a Max$-$2SE threshold (or Max$-$SE for more conservative post-onset detection) calibrated on training scores for automated alarming.

\titlespacing*{\section}{0pt}{2ex}{1ex}  % left, before, after
\section{Discussion and Conclusion} \label{sec:discussion}
\titlespacing*{\section}{0pt}{*3}{*1.5} 
This paper presented a label-free, offline-trainable fault detection system based on adversarial inverse reinforcement learning, evaluated on three industrial run-to-failure benchmarks. While our primary objective remains achieving the \textit{earliest possible detection}, AIRL was the only method to {validate these early warnings by achieving} non-saturated, consistent PDC (68--90\%) across all three datasets while respecting onset ground-truth timelines. Reconstruction-based baselines (AE, VAE, LSTM variants, USAD) saturate {indiscriminately}; TCAE and CTCAE suffer mode collapse on HUMS; ATF saturates on HUMS and XJTU-SY, and collapses on IMS under strict thresholds despite low FAR under Otsu. These results confirm that higher reconstruction capacity does not prevent the saturation-or-collapse failure inherent to run-to-failure prognostics. {AIRL's temporal reward modeling is the key differentiator.}

\textbf{Scope of the RL claim:} {Our argument is {precise} and deliberate: we are not claiming} reinforcement learning is universally superior to supervised or signal-processing methods for fault detection. Simple machine-learning and signal-processing baselines frequently match or outperform deep models in this domain~\cite{wang2023deep, neupane2024comparative, sawalhi2024helicopter}, and supervised methods remain the right tool for fault \textit{classification} once detection has flagged an anomaly~\cite{zhang2020deep,kibrete2024multi}. {Our claim is the {more specific} one:} \textit{if} RL is to be applied to MFD, the prevailing contextual-bandit formulation defeats its purpose. Genuine RL, whether online or offline, is distinguished by reasoning over \textit{state transitions} and credit-assignment through \textit{delayed reward}~\cite{barto2004reinforcement}; an episode that terminates after a single observation reduces to supervised learning. To operationalize sequential reasoning without physical control actions, we adopted a state-only imitation learning perspective~\cite{torabi2018generative}, framing the next observation as a proxy action. This recovers a reward function that measures the \textit{plausibility of the transition itself}, shifting focus from ``what does the signal look like?'' to ``how is the signal evolving?'' The contextual-bandit baseline (CTQN), which lacks this temporal reasoning, exhibited mode collapse on HUMS ({all files flagged Faulty,} FAR=100\%) and {pre-GT false alarms on IMS, empirically confirming the theoretical argument; its near-onset detection on XJTU-SY (File~455) does not generalize.}

\textbf{Contribution:} To our knowledge this is the first principled application of inverse RL to machinery fault detection, demonstrating that offline RL can be applied to industrial prognostic data without collapsing into supervised classification. The contribution is methodological: a transferable blueprint for using sequential RL on offline industrial sensor data, with an associated evaluation framework (timeliness, FAR, PDC, saturation flagging) that surfaces failure modes which standard metrics like AUROC hide.

\textbf{Limitations and future work:} Three limitations are worth naming explicitly. First, AIRL operates on single-channel scalar transitions $(x_t, x_{t+1}) \in \mathbb{R}^2$, which keeps the network compact and edge-deployable but discards cross-channel correlations that multi-sensor industrial systems routinely provide; multi-sensor fusion with shared discriminator architectures is a logical extension. Second, AIRL's performance is dataset-dependent: on IMS, the gradual nature of bearing degradation produces a compressed AIRL score range, which translates to a higher pre-GT FAR (88\%) than on HUMS (42\%) or XJTU-SY (68\%). While AIRL remains the only method maintaining non-saturated FAR across all three datasets, the IMS result indicates that gradual bearing degradation is the harder setting for transition-based reward learning, and motivates non-stationary or distribution-shift-aware threshold adaptation. Third, validation is currently limited to three public run-to-failure datasets; larger industrial corpora, online deployment trials, and uncertainty quantification on the learned reward function remain future work.

\appendix \label{sec:appendix}
\titlespacing*{\section}{0pt}{2ex}{1ex}
\section{Ablation Study: Impact of Sequential Reasoning}
\titlespacing*{\section}{0pt}{*3}{*1.5}

To validate the role of sequential modeling, we ablated the discount factor ($\gamma$), comparing AIRL ($\gamma=0.99$) against a myopic variant ($\gamma=0$) that reduces the agent to a contextual bandit. The sequential formulation offers three measurable advantages{, each targeting a distinct failure mode of the bandit formulation:}

\noindent\textbf{(i) Tighter healthy-manifold representation:} On HUMS2023, the sequential discriminator loss stabilized at $\approx 0.75$ versus $\approx 0.83$ for the myopic baseline, and the $p_{99}$ anomaly threshold was lower ($\tau \approx 0.319$ vs.\ $0.347$). Despite the stricter threshold, the sequential model flagged a higher total anomaly volume, indicating that the value function characterizes healthy behavior precisely enough that subtle fault indicators register with high confidence {rather than requiring large deviations}.

\noindent\textbf{(ii) Noise filtering via temporal credit:} On XJTU-SY, the myopic model triggered false alarms within the known healthy window ($<500$~min): File~277 under max-SE and File~261 under max-2SE. The sequential model filtered these fluctuations and delayed detection to File~533 (max-SE) and File~459 (max-2SE). {Without temporal credit assignment, the bandit formulation cannot distinguish transient instability from genuine degradation.}

\noindent\textbf{(iii) Stability across thresholds:} On IMS, the myopic baseline produced zero detections under strict thresholds (max, max-SE); its earliest detection (Nov~9) required loose adaptive thresholds (GPD) and collapsed under stricter calibration. The sequential model triggered on Nov~17, within the mid-November degradation window~\cite{berghout2021leveraging}, and maintained dense, stable detection across the full threshold hierarchy.

{Their cumulative effect explains why AIRL's full $\gamma=0.99$ configuration consistently outperforms the myopic ablation across all three datasets reported in the main results.}

\section*{Usage of Generative AI}
The GenAI models (and Copilot) were used only to fix the LaTeX equations and formatting issues, refine  sentences or words, fixing minor bugs in code. All the generated  suggestions were thoroughly verified, edited, and rewritten by the authors, and no AI text, figures, or code appears in the manuscript. The authors maintain full responsibility and accountability for all concepts \& methodology, data, analysis, and results \& conclusions, and this limited use of GenAI complies with ACM's authorship and transparency policies.

\bibliographystyle{ACM-Reference-Format}
\bibliography{zReferences}

@String{Computing = "Computing" }

@String{Springer = "Springer-Verlag" }

@book{sutton2020reinforcement,
  title={Reinforcement Learning: An Introduction},
  author={Sutton, Richard S. and Barto, Andrew G.},
  year={2020},
  edition={Second},
  series={Adaptive Computation and Machine Learning},
  publisher={MIT Press},
  address={Cambridge, Massachusetts London, England}
}

@article{neupane2024MFDReview,
  title={Data-driven machinery fault diagnosis: A comprehensive review},
  author={Neupane, Dhiraj and Bouadjenek, Mohamed Reda and Dazeley, Richard and Aryal, Sunil},
  journal={Neurocomputing},
  pages={129588},
  year={2025},
  publisher={Elsevier}
}

@article{ramirez2023semiSupervisedReview,
  title={Semi-supervised learning for industrial fault detection and diagnosis: A systemic review},
  author={Ram{\'\i}rez-Sanz, Jos{\'e} Miguel and Maestro-Prieto, Jose-Alberto and Arnaiz-Gonz{\'a}lez, {\'A}lvar and Bustillo, Andr{\'e}s},
  journal={ISA Transactions},
  year={2023},
  publisher={Elsevier}
}

@ARTICLE{9552620unsupervisedTransferReview,
  author={Zhao, Zhibin and Zhang, Qiyang and Yu, Xiaolei and Sun, Chuang and Wang, Shibin and Yan, Ruqiang and Chen, Xuefeng},
  journal={IEEE Transactions on Instrumentation and Measurement}, 
  title={Applications of Unsupervised Deep Transfer Learning to Intelligent Fault Diagnosis: A Survey and Comparative Study}, 
  year={2021},
  volume={70},
  number={},
  pages={1-28},
  doi={10.1109/TIM.2021.3116309}
}

@inproceedings{neupane2024machineryDS,
  title={Machinery Fault Detection using Advanced Machine Learning Techniques},
  author={Neupane, Dhiraj and Bouadjenek, Mohamed Reda and Dazeley, Richard and Aryal, Sunil},
  booktitle={PHM Society European Conference},
  volume={8},
  number={1},
  pages={4--4},
  year={2024}
}

@article{ding2019intelligent,
  title={Intelligent fault diagnosis for rotating machinery using deep Q-network based health state classification: A deep reinforcement learning approach},
  author={Ding, Yu and Ma, Liang and Ma, Jian and Suo, Mingliang and Tao, Laifa and Cheng, Yujie and Lu, Chen},
  journal={Advanced Engineering Informatics},
  volume={42},
  pages={100977},
  year={2019},
  publisher={Elsevier}
}

@article{qian2022development,
  title={Development of deep reinforcement learning-based fault diagnosis method for rotating machinery in nuclear power plants},
  author={Qian, Gensheng and Liu, Jingquan},
  journal={Progress in Nuclear Energy},
  volume={152},
  pages={104401},
  year={2022},
  publisher={Elsevier}
}

@article{nian2020review,
  title={A review on reinforcement learning: Introduction and applications in industrial process control},
  author={Nian, Rui and Liu, Jinfeng and Huang, Biao},
  journal={Computers \& Chemical Engineering},
  volume={139},
  pages={106886},
  year={2020},
  publisher={Elsevier}
}

@misc{wang2023helicopter,
  title={Helicopter main gearbox planet gear crack propagation test dataset},
  author={Wang, Wenyi and Blunt, David and Kappas, J},
  year={2023}
}

@article{sawalhi2024helicopter,
  title={Helicopter planet gear rim crack diagnosis and trending using cepstrum editing enhanced with deconvolution},
  author={Sawalhi, Nader and Wang, Wenyi and Blunt, David},
  journal={Sensors},
  volume={24},
  number={8},
  pages={2593},
  year={2024},
  publisher={MDPI}
}

@article{peeters2024fatigue,
  title={Fatigue crack detection in planetary gears: Insights from the HUMS2023 data challenge},
  author={Peeters, C{\'e}dric and Wang, Wenyi and Blunt, David and Verstraeten, Timothy and Helsen, Jan},
  journal={Mechanical Systems and Signal Processing},
  volume={212},
  pages={111292},
  year={2024},
  publisher={Elsevier}
}

@article{li2025novel,
  title={A novel reinforcement learning agent for rotating machinery fault diagnosis with data augmentation},
  author={Li, Zhenning and Jiang, Hongkai and Wang, Xin},
  journal={Reliability Engineering \& System Safety},
  volume={253},
  pages={110570},
  year={2025},
  publisher={Elsevier}
}

@inproceedings{neupane2024comparative,
  title={A Comparative Study of Semi-Supervised Anomaly Detection Methods for Machine Fault Detection},
  author={Neupane, Dhiraj and Bouadjenek, Mohamed Reda and Dazeley, Richard and Aryal, Sunil},
  booktitle={PHM Society European Conference},
  volume={8},
  number={1},
  pages={10--10},
  year={2024}
}

@article{li2025convolutional,
  title={A convolutional-transformer reinforcement learning agent for rotating machinery fault diagnosis},
  author={Li, Zhenning and Jiang, Hongkai and Dong, Yutong},
  journal={Expert Systems with Applications},
  volume={271},
  pages={126669},
  year={2025},
  publisher={Elsevier}
}

@inproceedings{ng2000algorithms,
  title={Algorithms for inverse reinforcement learning.},
  author={Ng, Andrew Y and Russell, Stuart and others},
  booktitle={Icml},
  volume={1},
  number={2},
  pages={2},
  year={2000}
}

@article{nunes2023challenges,
  title={Challenges in predictive maintenance--A review},
  author={Nunes, P and Santos, J and Rocha, E},
  journal={CIRP Journal of Manufacturing Science and Technology},
  volume={40},
  pages={53--67},
  year={2023},
  publisher={Elsevier}
}

@article{deshpande2025advances,
  title={Advances and applications in inverse reinforcement learning: a comprehensive review},
  author={Deshpande, Saurabh and Walambe, Rahee and Kotecha, Ketan and Selvachandran, Ganeshsree and Abraham, Ajith},
  journal={Neural Computing and Applications},
  pages={1--53},
  year={2025},
  publisher={Springer}
}

@article{saied2025development,
  title={Development of deep reinforcement learning-based fault diagnosis method for actuator faults in unmanned aerial vehicles},
  author={Saied, M and Tahan, N and Chreif, K and Francis, C and Noun, Z},
  journal={The Aeronautical Journal},
  pages={1--17},
  year={2025},
  publisher={Cambridge University Press}
}

@article{li2021deep,
  title={Deep reinforcement learning-based online domain adaptation method for fault diagnosis of rotating machinery},
  author={Li, Guoqiang and Wu, Jun and Deng, Chao and Xu, Xuebing and Shao, Xinyu},
  journal={IEEE/ASME Transactions on Mechatronics},
  volume={27},
  number={5},
  pages={2796--2805},
  year={2021},
  publisher={IEEE}
}

@article{yang2023new,
  title={A new intelligent fault diagnosis framework for rotating machinery based on deep transfer reinforcement learning},
  author={Yang, Daoguang and Karimi, Hamid Reza and Pawelczyk, Marek},
  journal={Control Engineering Practice},
  volume={134},
  pages={105475},
  year={2023},
  publisher={Elsevier}
}

@article{xiao2024research,
  title={Research on intelligent fault diagnosis for railway point machines using deep reinforcement learning},
  author={Xiao, Shuai and Feng, Qingsheng and Li, Xue and Li, Hong},
  journal={Transportation Safety and Environment},
  volume={6},
  number={4},
  pages={tdae007},
  year={2024},
  publisher={Oxford University Press}
}

@article{barto2004reinforcement,
  title={Reinforcement learning and its relationship to supervised learning},
  author={Barto, Andrew G and Dietterich, Thomas G},
  journal={Handbook of learning and approximate dynamic programming},
  volume={10},
  pages={9780470544785},
  year={2004},
  publisher={New York: Wiley-IEEE Press}
}

@article{wang2023deep,
  title={Is deep learning superior to traditional techniques in machine health monitoring applications},
  author={Wang, W and Vos, K and Taylor, J and Jenkins, C and Bala, B and Whitehead, L and Peng, Z},
  journal={The Aeronautical Journal},
  volume={127},
  number={1318},
  pages={2105--2117},
  year={2023},
  publisher={Cambridge University Press}
}

@article{zhang2020deep,
  title={Deep learning algorithms for bearing fault diagnostics—A comprehensive review},
  author={Zhang, Shen and Zhang, Shibo and Wang, Bingnan and Habetler, Thomas G},
  journal={IEEE access},
  volume={8},
  pages={29857--29881},
  year={2020},
  publisher={IEEE}
}

@article{kibrete2024multi,
  title={Multi-Sensor data fusion in intelligent fault diagnosis of rotating machines: A comprehensive review},
  author={Kibrete, Fasikaw and Woldemichael, Dereje Engida and Gebremedhen, Hailu Shimels},
  journal={Measurement},
  pages={114658},
  year={2024},
  publisher={Elsevier}
}

@article{torabi2018generative,
  title={Generative adversarial imitation from observation},
  author={Torabi, Faraz and Warnell, Garrett and Stone, Peter},
  journal={arXiv preprint arXiv:1807.06158},
  year={2018}
}

@inproceedings{torabi2019adversarial,
  title={Adversarial imitation learning from state-only demonstrations},
  author={Torabi, Faraz and Warnell, Garrett and Stone, Peter},
  booktitle={Proceedings of the 18th International Conference on Autonomous Agents and MultiAgent Systems},
  pages={2229--2231},
  year={2019}
}

@inproceedings{NEURIPS2024_f7faa46b,
 author = {Huang, Bo-Ruei and Yang, Chun-Kai and Lai, Chun-Mao and Wu, Dai-Jie and Sun, Shao-Hua},
 booktitle = {Advances in Neural Information Processing Systems},
 editor = {A. Globerson and L. Mackey and D. Belgrave and A. Fan and U. Paquet and J. Tomczak and C. Zhang},
 pages = {137190--137217},
 publisher = {Curran Associates, Inc.},
 title = {Diffusion Imitation from Observation},
 url = {https://proceedings.neurips.cc/paper_files/paper/2024/file/f7faa46b563c2e5343a728c85bace833-Paper-Conference.pdf},
 volume = {37},
 year = {2024}
}

@article{zhu2020off,
  title={Off-policy imitation learning from observations},
  author={Zhu, Zhuangdi and Lin, Kaixiang and Dai, Bo and Zhou, Jiayu},
  journal={Advances in neural information processing systems},
  volume={33},
  pages={12402--12413},
  year={2020}
}

@inproceedings{vaerenberg2025detecting,
  title={Detecting planet gear crack propagation using FRESH filters},
  author={Vaerenberg, Rik and Ricardo Mauricio, Alex and Gryllias, Konstantinos},
  booktitle={14th Defence Science \& Technology (DST) International Conference on Health and Usage Monitoring HUMS2025 Proceedings},
  year={2025}
}

@article{sacerdoti2023comparison,
  title={A comparison of signal analysis techniques for the diagnostics of the IMS rolling element bearing dataset},
  author={Sacerdoti, Diletta and Strozzi, Matteo and Secchi, Cristian},
  journal={Applied Sciences},
  volume={13},
  number={10},
  pages={5977},
  year={2023},
  publisher={MDPI}
}

@article{scholkopf2001estimating,
  title={Support vector method for novelty detection},
  author={Sch{\"o}lkopf, Bernhard and Williamson, Robert C and Smola, Alex and Shawe-Taylor, John and Platt, John},
  journal={Advances in neural information processing systems},
  volume={12},
  year={1999}
}

@article{malhotra2016lstm,
  title={LSTM-based encoder-decoder for multi-sensor anomaly detection},
  author={Malhotra, Pankaj and Ramakrishnan, Anusha and Anand, Gaurangi and Vig, Lovekesh and Agarwal, Puneet and Shroff, Gautam},
  journal={arXiv preprint arXiv:1607.00148},
  year={2016}
}

@article{mao2024harmony,
  title={Harmony better than uniformity: A new pre-training anomaly detection method with tensor domain adaptation for early fault evaluation},
  author={Mao, Wentao and Chen, Zongtao and Zhang, Yanna and Zhong, Zhidan},
  journal={Engineering Applications of Artificial Intelligence},
  volume={127},
  pages={107427},
  year={2024},
  publisher={Elsevier}
}

@article{wang2020xjtusy,
  author  = {Wang, Biao and Lei, Yaguo and Li, Naipeng and Li, Ningbo},
  title   = {A Hybrid Prognostics Approach for Estimating Remaining Useful Life of Rolling Element Bearings},
  journal = {IEEE Transactions on Reliability},
  year    = {2020},
  volume  = {69},
  number  = {1},
  pages   = {401--412},
  doi     = {10.1109/TR.2018.2882682}
}

@inproceedings{fu2018airl,
  author    = {Fu, Justin and Luo, Katie and Levine, Sergey},
  title     = {Learning Robust Rewards with Adversarial Inverse Reinforcement Learning},
  booktitle = {International Conference on Learning Representations (ICLR)},
  year      = {2018},
  url       = {https://openreview.net/forum?id=rkHywl-A-}
}

@inproceedings{liu2008isolation,
  title={Isolation Forest},
  author={Liu, Fei Tony and Ting, Kai Ming and Zhou, Zhi-Hua},
  booktitle={2008 Eighth IEEE International Conference on Data Mining},
  pages={413--422},
  year={2008},
  organization={IEEE}
}

@inproceedings{zhou2017anomaly,
  title={Anomaly detection with robust deep autoencoders},
  author={Zhou, Chong and Paffenroth, Randy C},
  booktitle={Proceedings of the 23rd ACM SIGKDD International Conference on Knowledge Discovery and Data Mining},
  pages={665--674},
  year={2017}
}

@article{kingma2013auto,
  title={Auto-encoding variational Bayes},
  author={Kingma, Diederik P and Welling, Max},
  journal={arXiv preprint arXiv:1312.6114},
  year={2013}
}

@article{chung2015vrnn,
  title={A recurrent latent variable model for sequential data},
  author={Chung, Junyoung and Kastner, Kyle and Dinh, Laurent and Goel, Kratarth and Courville, Aaron C and Bengio, Yoshua},
  journal={Advances in neural information processing systems},
  volume={28},
  year={2015}
}

@article{otsu1979threshold,
  title={A threshold selection method from gray-level histograms},
  author={Otsu, Nobuyuki and others},
  journal={Automatica},
  volume={11},
  number={285-296},
  pages={23--27},
  year={1975}
}

@inproceedings{macqueen1967kmeans,
  title={Some methods of classification and analysis of multivariate observations},
  author={McQueen, James B},
  booktitle={Proc. of 5th Berkeley Symposium on Math. Stat. and Prob.},
  pages={281--297},
  year={1967}
}

@article{pickands1975statistical,
  title={Statistical inference using extreme order statistics},
  author={Pickands III, James},
  journal={the Annals of Statistics},
  pages={119--131},
  year={1975},
  publisher={JSTOR}
}

@article{romo2024median,
  title={Median absolute deviation for BGP anomaly detection},
  author={Romo-Chavero, Maria Andrea and Cantoral-Ceballos, Jose Antonio and P{\'e}rez-D{\'\i}az, Jesus Arturo and Martinez-Cagnazzo, Carlos},
  journal={Future Internet},
  volume={16},
  number={5},
  pages={146},
  year={2024},
  publisher={MDPI}
}

@article{umsonst2022finite,
  title={Finite sample guarantees for quantile estimation: An application to detector threshold tuning},
  author={Umsonst, David and Ruths, Justin and Sandberg, Henrik},
  journal={IEEE Transactions on Control Systems Technology},
  volume={31},
  number={2},
  pages={921--928},
  year={2022},
  publisher={IEEE}
}

@article{yang2025agent,
  title={Agent-based dynamic thresholding for adaptive anomaly detection using reinforcement learning},
  author={Yang, Xue and Howley, Enda and Schukat, Michael},
  journal={Neural Computing and Applications},
  volume={37},
  number={23},
  pages={18775--18791},
  year={2025},
  publisher={Springer}
}

@article{mao2020robust,
  title={Robust detection of bearing early fault based on deep transfer learning},
  author={Mao, Wentao and Zhang, Di and Tian, Siyu and Tang, Jiamei},
  journal={Electronics},
  volume={9},
  number={2},
  pages={323},
  year={2020},
  publisher={MDPI}
}

@inproceedings{chen2020simple,
  title={A simple framework for contrastive learning of visual representations},
  author={Chen, Ting and Kornblith, Simon and Norouzi, Mohammad and Hinton, Geoffrey},
  booktitle={International conference on machine learning},
  pages={1597--1607},
  year={2020},
  organization={PmLR}
}

@article{berghout2021leveraging,
  title={Leveraging label information in a knowledge-driven approach for rolling-element bearings remaining useful life prediction},
  author={Berghout, Tarek and Benbouzid, Mohamed and Mouss, Le{\"\i}la-Hayet},
  journal={Energies},
  volume={14},
  number={8},
  pages={2163},
  year={2021},
  publisher={MDPI}
}

@inproceedings{audibert2020usad,
   title={{USAD}: Unsupervised anomaly detection on multivariate time series},
   author={Audibert, Julien and Michiardi, Pietro and Guyard, Fr{\'e}d{\'e}ric and Marti, S{\'e}bastien and Zuluaga, Maria A},
   booktitle={Proceedings of the 26th ACM SIGKDD International Conference on Knowledge Discovery \& Data Mining},
   pages={3395--3404},
   year={2020}
}

@inproceedings{xu2022anomaly,
   title={Anomaly Transformer: Time series anomaly detection with association discrepancy},
   author={Xu, Jiehui and Wu, Haixu and Wang, Jianmin and Long, Mingsheng},
   booktitle={International Conference on Learning Representations (ICLR)},
   year={2022}
}

@article{lei2018machinery,
  title={Machinery health prognostics: A systematic review from data acquisition to RUL prediction},
  author={Lei, Yaguo and Li, Naipeng and Guo, Liang and Li, Ningbo and Yan, Tao and Lin, Jing},
  journal={Mechanical systems and signal processing},
  volume={104},
  pages={799--834},
  year={2018},
  publisher={Elsevier}
}

@article{li2018remaining,
  title={Remaining useful life estimation in prognostics using deep convolution neural networks},
  author={Li, Xiang and Ding, Qian and Sun, Jian-Qiao},
  journal={Reliability Engineering \& System Safety},
  volume={172},
  pages={1--11},
  year={2018},
  publisher={Elsevier}
}

@inproceedings{neupanelearning,
  title={Learning Rewards, Not Labels: Adversarial Inverse Reinforcement Learning for Machinery Fault Detection},
  author={Neupane, Dhiraj and Dazeley, Richard and Bouadjenek, Mohamed Reda and Aryal, Sunil},
  booktitle={Proceedings of the 25th International Conference on Autonomous Agents and Multiagent Systems (AAMAS)},
  year={2026},
  doi={10.65109/AXYX4522}
}

@article{bai2018empirical,
  title={An empirical evaluation of generic convolutional and recurrent networks for sequence modeling},
  author={Bai, Shaojie and Kolter, J Zico and Koltun, Vladlen},
  journal={arXiv preprint arXiv:1803.01271},
  year={2018}
}

\end{document}